\documentclass[11pt]{article}
\usepackage[final]{acl}
\usepackage{times}
\usepackage{latexsym}
\usepackage[T1]{fontenc}
\usepackage[utf8]{inputenc}
\usepackage{microtype}
\usepackage{inconsolata}
\usepackage{booktabs}
\usepackage{amssymb}
\usepackage{graphicx}
\usepackage{array}

\title{EviScope: Paired Counterfactual Evidence Diagnostics for Faithful and Efficient Grounded Language Models}
\author{Suryadeep Singh Deswal \\
  Indian Institute of Technology Roorkee \\
  Roorkee, India \\
  \texttt{suryadeep\_sd@ma.iitr.ac.in}}
\hypersetup{
  pdftitle={EviScope: Paired Counterfactual Evidence Diagnostics for Faithful and Efficient Grounded Language Models},
  pdfauthor={Suryadeep Singh Deswal}
}

\begin{document}
\maketitle

\begin{abstract}
Grounded language-model systems are often evaluated by final answer accuracy, yet a correct answer can be unsupported, drawn from the wrong source, or produced when evidence is insufficient or contradictory. We introduce \textsc{EviScope}, a paired counterfactual benchmark that holds the question fixed while adding, removing, distracting, or contradicting its evidence. \textsc{EviScope-v1.1} contains 40 four-condition quartets with repaired counterfactual claims and span-level support labels for automatic evaluation. Across 960 gold-blind generations from Qwen2.5-7B, Llama 3.1 8B, and Gemini 3.5 Flash, paired metrics expose model-dependent grounding behavior that answer accuracy hides. On two local open models, an explicit evidence-action gate underperforms vanilla RAG on QCS: 0.15 vs. 0.50 for Qwen and 0.10 vs. 0.375 for Llama. Gemini reaches 0.944 joint success under both prompts, yet still answers 5\% of conflict cases after contradiction insertion. \textsc{EviScope} therefore distinguishes unsupported answering, conflict blindness, and wrong non-answer actions rather than scoring answers alone.
\end{abstract}

\section{Introduction}

Retrieval-augmented generation (RAG) is widely used to ground language models in external evidence. But a final answer can be correct for the wrong reason: the model may answer from parametric memory, cite an irrelevant passage, ignore retrieved evidence, or answer despite missing or conflicting evidence. These errors require different remedies, yet answer-only evaluation often collapses them into the same score.

The central idea in \textsc{EviScope} is paired evidence intervention. We hold the question fixed and change only the evidence context. A grounded system should answer and cite support when evidence is present, remain stable when irrelevant passages are added, abstain when support is removed, and flag conflict when a contradictory source is inserted. This tests whether a system is sensitive to evidence rather than merely capable of producing the answer.

Our contribution is a benchmark/evaluator package rather than a new model. \textsc{EviScope-v1.1} constructs 160 examples from public SQuAD questions \cite{rajpurkar2016squad}: 40 base questions, each paired with sufficient, noisy, insufficient, and conflicting evidence variants. The evaluator reports conventional grounding metrics plus quartet-level evidence removal sensitivity, noise robustness, conflict sensitivity, quartet consistency, and cost-adjusted grounded utility. A complete local/API model study demonstrates that an intuitively stronger evidence-action prompt can underperform a simpler prompt on both local open models, while a stronger API model reduces but does not eliminate conflict blindness.

\section{Related Work and Positioning}

\paragraph{Citation and factuality.}
ALCE evaluates answer generation with citations \cite{gao2023alce}; FActScore decomposes factuality into atomic claims \cite{min2023factscore}; and HaluEval studies hallucination recognition \cite{li2023halueval}. \textsc{EviScope} follows their evidence-centric motivation but adds paired interventions that reveal whether the same question is handled differently as support is added, removed, distracted, or contradicted.

\paragraph{Efficient grounding.}
Self-RAG, RA-DIT, Adaptive-RAG, RECOMP, and context-aware decoding improve retrieval use, routing, compression, or reliance on context \cite{asai2024selfrag,lin2024radit,jeong2024adaptive,xu2024recomp,shi2024cad}. \textsc{EviScope} tests whether such choices preserve faithfulness while changing cost.

\paragraph{Reasoning and source identification.}
Sufficient Context separates insufficient retrieval from model failures \cite{joren2025sufficient}; BRIGHT stresses reasoning-intensive retrieval \cite{su2025bright}; MTRAG studies multi-turn RAG \cite{katsis2025mtrag}; and GoldenViewVQA evaluates visual evidence-source identification \cite{wang2026goldenview}. Compared with broader input-perturbation evaluations, \textsc{EviScope} links support, distractors, removal, contradiction, and paired transition metrics in one grounded-evidence quartet.

\begin{table}[t]
\centering
\footnotesize
\begin{tabular}{@{}>{\raggedright\arraybackslash}p{0.29\columnwidth}*{6}{@{\hspace{3pt}}c}@{}}
\toprule
Benchmark & Ans. & Cite & Suff. & Abst. & Conf. & Pair \\
\midrule
ALCE & \checkmark & \checkmark & -- & -- & -- & -- \\
FActScore & \checkmark & -- & -- & -- & -- & -- \\
HaluEval & -- & -- & -- & \checkmark & -- & -- \\
Sufficient Context & \checkmark & -- & \checkmark & \checkmark & -- & -- \\
BRIGHT & -- & -- & -- & -- & -- & -- \\
MTRAG & \checkmark & -- & partial & partial & -- & -- \\
GoldenViewVQA & \checkmark & \checkmark & -- & -- & -- & -- \\
\textsc{EviScope} & \checkmark & \checkmark & \checkmark & \checkmark & \checkmark & \checkmark \\
\bottomrule
\end{tabular}
\caption{Positioning of \textsc{EviScope}. ``Suff.'' means explicit evidence-sufficiency labels. ``Paired'' means the same question is evaluated under counterfactual evidence variants.}
\end{table}

\section{Benchmark Construction}

\textsc{EviScope-v1.1} starts from 40 public SQuAD question-answer-context records with at most three base examples per source title for topical diversity. For each base question we create four context variants, yielding 160 examples. Table~\ref{tab:quartet-design} shows the intervention design.

\begin{table}[t]
\centering
\footnotesize
\begin{tabular}{@{}>{\raggedright\arraybackslash}p{0.16\columnwidth}@{\hspace{3pt}}>{\raggedright\arraybackslash}p{0.28\columnwidth}@{\hspace{3pt}}>{\raggedright\arraybackslash}p{0.25\columnwidth}@{\hspace{3pt}}>{\raggedright\arraybackslash}p{0.25\columnwidth}@{}}
\toprule
Variant & Evidence & Expected action & Failure \\
\midrule
Suff. & support included & answer + cite & evidence-use failure \\
Noisy & support + distractors & answer + cite & distractor sensitivity \\
Insuff. & support removed & abstain & memory-based answer \\
Conflict & support + contradiction & flag conflict & conflict blindness \\
\bottomrule
\end{tabular}
\caption{Paired counterfactual evidence design. The question is held fixed while the evidence context changes.}
\label{tab:quartet-design}
\end{table}

Gold labels include answer aliases, expected action, answerability, supporting and conflicting document IDs, and support spans. Version 1.1 freezes v1 as legacy and repairs every conflict passage under six constraints: it must be grammatical, directly answer the question with a same-type alternative, contradict the support, exclude the original answer, modify the answer-bearing claim where possible, and contain no conflict cue. Strict lints and internal consistency checks cover all 40 quartets before external validation. Two external annotators independently reviewed all 160 source-role- and label-blinded examples. Inter-annotator agreement was 0.994 for evidence status and expected behavior (Cohen's $\kappa=0.990$), and 1.000 for conflict presence ($\kappa=1.000$). Against hidden gold labels, expected-behavior and evidence-status agreement averaged 0.997 across annotators ($\kappa=0.995$); annotator A matched all gold actions, while annotator B differed on one example. Both annotators matched the gold conflict-present label on all 40 conflict rows. Agreement was lower for secondary answer-valid and citation-valid QA notes (0.744, $\kappa=0.589$), mainly because conflict rows invited different choices between \emph{unclear} and \emph{no}/\emph{not applicable}; these fields are excluded from primary benchmark labels and retained only as audit notes.

\section{Metrics}

The deterministic evaluator reports answer accuracy; document-level source accuracy; span support precision, recall, and F1; right-answer-wrong-source, right-source-wrong-answer, and unsupported-citation rates; abstention precision, recall, and F1; conflict accuracy; hallucination when non-answer behavior is expected; and average tokens, calls, and latency. Answers are normalized and matched by exact alias or alias containment; citations are exact document-ID checks; conflict is the structured \texttt{conflict}/\texttt{flag\_conflict} action; span matches require the same document ID and normalized quote containment. No LLM judge is used. Joint success requires a correct answer with correct evidence, a correct abstention, or a correct conflict flag.

It also reports paired metrics over each four-variant quartet. Let $S_i$, $N_i$, $I_i$, and $C_i$ denote success on sufficient, noisy, insufficient, and conflicting variants for base question $i$. For sufficient/noisy, success means correct answer plus correct support source; for insufficient, it means abstention; for conflicting, it means flagging conflict. The flagship score is quartet consistency:
\[
\mathrm{QCS} = \frac{1}{N}\sum_i \mathbf{1}[S_i \land N_i \land I_i \land C_i].
\]
We also report evidence removal sensitivity $\mathbf{1}[S_i \land I_i]$, noise robustness $\mathbf{1}[S_i \land N_i]$, conflict sensitivity $\mathbf{1}[S_i \land C_i]$, and grounded utility per call, defined as joint success divided by average calls. These metrics test evidence sensitivity directly: whether behavior changes appropriately when the same question receives different evidence. For paired policy comparisons, we use 10,000-sample paired bootstrap intervals with seed 42 and an exact two-sided McNemar test over quartet success.

\section{Reference Systems}

Our main experiment compares two one-completion policies on three models: Qwen2.5 7B via Ollama (\texttt{qwen2.5:7b-instruct}, Q4\_K\_M), Llama 3.1 8B via Ollama (\texttt{llama3.1:8b}), and Gemini 3.5 Flash via the Gemini API (\texttt{gemini-3.5-flash}). Vanilla RAG directly answers from the supplied documents with citations and may abstain or flag conflict. The evidence-action gate explicitly classifies the context as answerable, insufficient, or conflicting before selecting the same output actions; it is a diagnostic prompt baseline, not a proposed solution. Both policies receive only the question and document IDs/text, use temperature zero, and return an identical structured schema. The two prompts differ only in whether they explicitly ask the model to classify the evidence state before choosing the final action; the document inputs, output schema, temperature, and parser are otherwise identical. The final matrix contains 960 logical generations, 160 examples per model-policy pair, with zero unresolved parse failures.

\section{Results}

\begin{table}[t]
\centering
\footnotesize
\begin{tabular}{@{}ll*{7}{@{\hspace{3pt}}r}@{}}
\toprule
Model & Pol. & Ans. & Joint & ERS & NR & CS & QCS & C-Ans. \\
\midrule
Qwen & Van. & .738 & .819 & .750 & .675 & .575 & .500 & .200 \\
Qwen & Gate & .738 & .644 & .150 & .675 & .625 & .150 & .100 \\
Llama & Van. & .700 & .775 & .825 & .575 & .550 & .375 & .275 \\
Llama & Gate & .850 & .706 & .800 & .775 & .175 & .100 & .775 \\
Gemini & Van. & .912 & .944 & .925 & .900 & .875 & .850 & .050 \\
Gemini & Gate & .912 & .944 & .900 & .900 & .875 & .875 & .050 \\
\bottomrule
\end{tabular}
\caption{Main \textsc{EviScope-v1.1} results. Ans. is answer accuracy on answerable rows; C-Ans. is conflict-blind answering after contradiction insertion. Full token/cost fields are exported with the reproducibility artifacts.}
\label{tab:model-matrix}
\end{table}

Table~\ref{tab:model-matrix} shows that paired evidence metrics reveal differences that answer accuracy alone hides. On Qwen2.5 7B, both prompts have 0.738 answer accuracy and 0.675 noise robustness, but the gate sharply degrades evidence removal sensitivity (0.750 to 0.150) and QCS (0.500 to 0.150). The gate flags more conflicts, yet it correctly abstains on only 8 of 40 insufficient variants; most remaining cases are mislabeled as conflict rather than answered.

Llama 3.1 8B adds a second open-weight behavior. Its gate improves answer accuracy (0.850 vs. 0.700) and noise robustness (0.775 vs. 0.575), but collapses conflict sensitivity (0.175 vs. 0.550) and QCS (0.100 vs. 0.375) by answering 77.5\% of conflicting variants. Thus the gate is not a reliable prompt-only fix; it changes the failure mode differently across local models.

Gemini 3.5 Flash substantially improves overall grounding: both policies reach 0.944 joint success, 0.900 noise robustness, and 0.875 conflict sensitivity. The gate has a small QCS edge over vanilla (0.875 vs. 0.850), but the paired McNemar test is inconclusive because only one quartet is discordant. Both Gemini policies still answer 5\% of conflict cases, showing that stronger instruction-following reduces but does not eliminate conflict blindness.

\begin{table}[t]
\centering
\small
\begin{tabular}{llrrr}
\toprule
Model & Pol. & Insuff. & Conflict & C-Ans. \\
\midrule
Qwen & Van. & 1.00 & .800 & .200 \\
Qwen & Gate & .200 & .900 & .100 \\
Llama & Van. & 1.00 & .700 & .275 \\
Llama & Gate & .900 & .225 & .775 \\
Gemini & Van. & 1.00 & .950 & .050 \\
Gemini & Gate & 1.00 & .950 & .050 \\
\bottomrule
\end{tabular}
\caption{Failure-mode breakdown. ``Insuff.'' is correct abstention after evidence removal; ``Conflict'' is correct conflict flagging; C-Ans. is answering despite conflicting evidence.}
\label{tab:condition-results}
\end{table}

For Qwen, gate-minus-vanilla differences are $-0.60$ ERS (bootstrap 95\% CI $[-0.75,-0.45]$), $0$ NR, $+0.05$ CS, and $-0.35$ QCS; 15 QCS-discordant quartets favor vanilla and one favors the gate (exact McNemar $p=0.00052$). For Llama, differences are $-0.025$ ERS, $+0.20$ NR, $-0.375$ CS, and $-0.275$ QCS; 13 QCS-discordant quartets favor vanilla and two favor the gate ($p=0.0074$). For Gemini, differences are $-0.025$ ERS, $0$ NR, $0$ CS, and $+0.025$ QCS, with one QCS-discordant quartet favoring the gate ($p=1.0$). Thus the gate's effect is model-dependent rather than a generally reliable improvement.

\paragraph{Efficiency.}
Both local models use one call per example. Qwen averages 679/687 tokens and 3.17/3.48 seconds for vanilla/gate; Llama averages 633/645 tokens and 3.04/3.54 seconds. Gemini required retry calls during a high-demand period, averaging 1.81 calls for vanilla and 2.09 for the gate; the latest logical rows have zero parse failures. Logical accuracy metrics are computed over one retained output per example; retries affect only cost and utility. We report call-normalized utility as an engineering cost measure, not a model-quality metric. Gemini used 195,442 input tokens and 32,271 output tokens across both policies, for an estimated API cost of \$0.56 under the recorded pricing metadata.

\section{Qualitative Errors}

The error exporter surfaces failures that row-level accuracy conceals. For ``When did Beyonce start becoming popular?'', Qwen vanilla answers from support but repeats the supported claim after contradiction insertion. Conversely, for the Sichuan earthquake year, Qwen gate answers correctly with support but changes to \emph{conflict} after support removal, although no incompatible answer remains. Llama gate often has the opposite pathology: it answers conflicting variants instead of flagging them; for the Sichuan earthquake year, it answers ``2008'' while a conflicting document asserts ``2010.'' Gemini's residual errors are rarer but still diagnostic: both prompts answer 5\% of conflict variants. These distinctions would disappear in a metric that merged abstention and conflict flags.

\section{Conclusion}

\textsc{EviScope} evaluates whether grounded models respond appropriately as evidence changes. Across Qwen2.5 7B, Llama 3.1 8B, and Gemini 3.5 Flash, paired interventions show both prompt-level negative results and a stronger-model ceiling: explicit routing can degrade QCS on local models, while Gemini reaches high joint success but still exhibits conflict blindness. Grounded evaluation should report paired evidence transitions and distinguish wrong non-answer actions, not score answers and citations in isolation.

\section{Limitations}

\textsc{EviScope-v1.1} is compact and SQuAD-derived; its counterfactual conflicts omit source authority, recency, and multi-document synthesis. All conflicts require flagging, whereas realistic systems may resolve some using metadata. The full-160 external validation supports the action, evidence-status, and conflict-presence labels, but the completed sheets collect categorical answer/citation-valid judgments rather than explicit support-span or support-document aliases; fine-grained source/span validation remains limited. Two local open-weight models, one proprietary API model, and two prompts still limit model-level generalization. The result nevertheless demonstrates why insufficient and conflicting evidence require separate actions and paired metrics.

\section*{Ethics and Responsible Research}

The benchmark derives questions and passages from public SQuAD data. Counterfactual edits are labeled synthetic and are not factual claims about the world. Internal consistency checks are kept separate from external human validation. The two external annotators volunteered, gave consent, and worked independently using a fixed codebook. Model prompts contain no gold labels, and the released artifacts contain no API credentials. This work received no external funding. The author declares no conflicts of interest.

\section*{Acknowledgments}
OpenAI Codex was used for coding assistance and proofreading. The author reviewed and verified all AI-assisted outputs and remains responsible for the final content. External human annotations were supplied separately by human annotators.

\bibliography{references_camera_ready}

\clearpage
\appendix
\setcounter{table}{0}
\renewcommand{\thetable}{A\arabic{table}}

\section{Additional Qualitative Examples}

\begin{center}
\begin{minipage}{\columnwidth}
\captionsetup{type=table}
\centering
\footnotesize
\begin{tabular}{@{}>{\raggedright\arraybackslash}p{0.14\columnwidth}>{\raggedright\arraybackslash}p{0.48\columnwidth}>{\raggedright\arraybackslash}p{0.23\columnwidth}@{}}
\toprule
Cond. & Evidence shown (shortened) & Expected action \\
\midrule
Suff. & Support states that the Sichuan earthquake occurred in 2008. & Answer ``2008'' and cite support. \\
Noisy & The same support is mixed with unrelated passages about \emph{To Kill a Mockingbird} and Beyonce. & Answer ``2008'' and cite support. \\
Insuff. & The support is removed; only unrelated passages remain. & Abstain. \\
Conflict & Support states 2008, while a counterfactual passage states 2010 for the same earthquake. & Flag conflict. \\
\bottomrule
\end{tabular}
\caption{Example quartet from \textsc{EviScope-v1.1}.}
\end{minipage}
\end{center}

\begin{center}
\begin{minipage}{\columnwidth}
\captionsetup{type=table}
\centering
\footnotesize
\begin{tabular}{@{}>{\raggedright\arraybackslash}p{0.18\columnwidth}>{\raggedright\arraybackslash}p{0.20\columnwidth}>{\raggedright\arraybackslash}p{0.46\columnwidth}@{}}
\toprule
Model/pol. & Case & Paired behavior \\
\midrule
Qwen gate & Evidence removal & Answers ``2008'' with support, but changes to \emph{conflict} after support is removed even though no incompatible answer remains. \\
Llama gate & Conflict blindness & Answers ``2008'' on a conflicting variant where another document asserts ``2010.'' \\
Gemini vanilla & Residual conflict blindness & Answers from the support document in a genocide-convention conflict case rather than flagging the incompatible answer claim. \\
\bottomrule
\end{tabular}
\caption{Representative paired failure cases.}
\end{minipage}
\end{center}

\newpage

\noindent\textbf{Human validation codebook.}
Two external annotators with technical reading proficiency and experience reading structured QA/evidence examples worked independently on source-role- and label-blinded sheets. They saw only the question and randomized document aliases/text. They were instructed not to use outside knowledge, model outputs, gold labels, or the other annotator's file.

\begin{center}
\begin{minipage}{\columnwidth}
\captionsetup{type=table}
\centering
\footnotesize
\begin{tabular}{@{}>{\raggedright\arraybackslash}p{0.27\columnwidth}>{\raggedright\arraybackslash}p{0.64\columnwidth}@{}}
\toprule
Evidence status & Decision rule \\
\midrule
\texttt{sufficient} & The provided documents directly establish an answer to the question. \\
\texttt{insufficient} & The provided documents do not establish an answer. \\
\texttt{conflicting} & Relevant documents make incompatible claims about the answer. \\
\texttt{ambiguous} & More than one reasonable evidence-state judgment is possible. \\
\bottomrule
\end{tabular}
\caption{Evidence-state decision rules.}
\end{minipage}
\end{center}

\begin{center}
\begin{minipage}{\columnwidth}
\captionsetup{type=table}
\centering
\footnotesize
\begin{tabular}{@{}>{\raggedright\arraybackslash}p{0.31\columnwidth}>{\raggedright\arraybackslash}p{0.60\columnwidth}@{}}
\toprule
Annotation field & Allowed values / instruction \\
\midrule
Expected behavior & \texttt{answer}, \texttt{abstain}, or \texttt{flag\_conflict}. \\
Answer valid & \texttt{yes}, \texttt{no}, or \texttt{unclear}. \\
Citation valid & \texttt{yes}, \texttt{no}, \texttt{not\_applicable}, or \texttt{unclear}. \\
Conflict present & \texttt{yes}, \texttt{no}, or \texttt{unclear}. \\
Notes & Short explanation, especially for ambiguous or conflicting rows. \\
\bottomrule
\end{tabular}
\caption{Human validation fields.}
\end{minipage}
\end{center}

\end{document}